\documentclass[9pt,twocolumn,twoside]{osajnl}

\journal{New} 

\setboolean{shortarticle}{false}

\ifthenelse{\boolean{shortarticle}}{\colorlet{color2}{color2b}}{\colorlet{color2}{color2}} 

\title{Using machine learning to create high-efficiency freeform illumination design tools}

\author[1]{Caleb Gannon}
\author[1,*]{Rongguang Liang}

\affil[1]{College of Optical Sciences, University of Arizona, 1630 East University Boulevard, Tucson, Arizona 85721, USA}

\affil[*]{Corresponding author: rliang@optics.arizona.edu}

 \dates{Compiled \today}

\begin{abstract}
We present a method for improving the efficiency and user experience of freeform illumination design with machine learning. By utilizing orthogonal polynomials to interface with artificial neural networks, we are able to generalize relationships between freeform surface shapes and design parameters. Then, by training the network to generalize the relationship between high-level design goals and final performance, we were able to transform what is traditionally a difficult and computationally intensive problem into a compact, user friendly form. The potential of the proposed method is demonstrated through the design of uniform square patterns from off-axis positions and rectangular patterns of tuneable aspect ratios and distances from the target. 
\end{abstract}

\setboolean{displaycopyright}{false}

\begin{document}

\maketitle
\thispagestyle{fancy}

\ifthenelse{\boolean{shortarticle}}{\ifthenelse{\boolean{singlecolumn}}{\abscontentformatted}{\abscontent}}{}

\section{Introduction}
The freeform illumination design problem is an extremely complex inverse problem where designers create optical surface(s) to produce a desired light distribution. In general this process requires calculating a mapping relationship between the source and target energy profiles and an optical surface to enforce this mapping via refraction or reflection. Simultaneous solutions for both of these parameters often require numerically solving a nonlinear, second order partial differential equation of the Monge-Ampere type \cite{Ries:02,wu2013freeform,brix2015designing,de2017numerical} or the calculation of SMS curves \cite{gimenez2004simultaneous}. While these methods are powerful and often able to produce exact solutions, they typically require a deep mathematical understanding and/or the specification of complicated boundary conditions. There have also been attempts to simplify this process and solve each task separately, for example \cite{gannon2017ray,extrinsic, Bauerle:12,fournier2010fast,feng2016freeform,Ma:15,bosel2016ray}, and although powerful, calculating the mapping separate from the surface construction introduces its own complexities such as ensuring the resulting surfaces fulfill the integrability condition. Fundamentally, freeform illumination design is a challenging task typically requiring years of expertise and relatively large amounts of computation time (typically minutes to hours for a single design). 

In this paper, we propose a method to help overcome both of these hurdles by using machine learning to simplify the design process. By teaching an artificial neural network the direct relationship between surface shape and desired performance we are able to bypass the complex intermediate calculations which are typically required to establish such a relationship. The end result is an efficient functional representation which accepts design parameters and outputs a completed optical surface, offering a dramatic speed improvement and simplified user experience.

\section{Artificial Intelligence}

\begin{figure}[!htb]
\centering
\includegraphics[width=\linewidth]{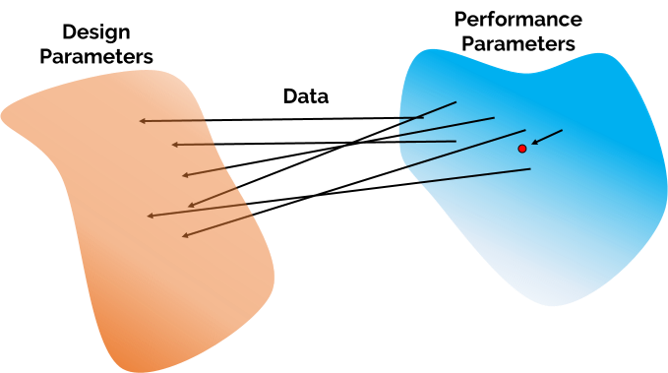}
\caption{An abstract representation of data created as a correspondence between the performance and design parameter spaces. The marked point denotes a desired performance characteristic which is similar to previous designs.}
\label{fig:AbstractSpaces}
\end{figure}

The concept of artificial intelligence discussed in this paper is relatively abstract, and is intended to describe an approach which is capable of building upon past experiences to improve future performance. In the case of the design process this means that after seeing existing designs, an intelligent agent should be able to use this knowledge to generate similar designs it had never seen before. To think mathematically how this can be done, we turn to Fig. \ref{fig:AbstractSpaces}. When designs are generated, they create a discrete mapping from one point in the performance parameter space to the design parameter space. Here the performance space corresponds to whatever representation the designer has chosen to describe the functionality of the system (for example irradiance on a target, an intensity distribution, energy efficiency, etc.) and the design space is a representation of the variables in the system design, such as lens shape, material, coatings etc. 

In Fig. \ref{fig:AbstractSpaces}, we marked a point in the performance space to denote a future design we would like to generate. Although it is relatively near to designs created in the past (denoted as data in the figure), there is no clear way to use this information to our benefit. Traditionally, an entirely new design would have to be generated every time we wanted to explore a new point in this space no matter how many similar designs were done previously. This can lead to many repeated calculations and wasted computational efforts.

As an alternative to this, rather than discarding designs after they are made we can present them to a learning agent that generates a continuous interpretation of the mapping relationship between these two spaces. Our agent acts as a universal function approximator \cite{HORNIK1989359}, and is intelligent in the sense that it can adapt this function to explain various input-output relationships. Since the agent's representation of this mapping relationship is a continuous function, with this approach we can now generate our desired design without issue. 

\begin{figure}[!htb]
\centering
\includegraphics[width=\linewidth]{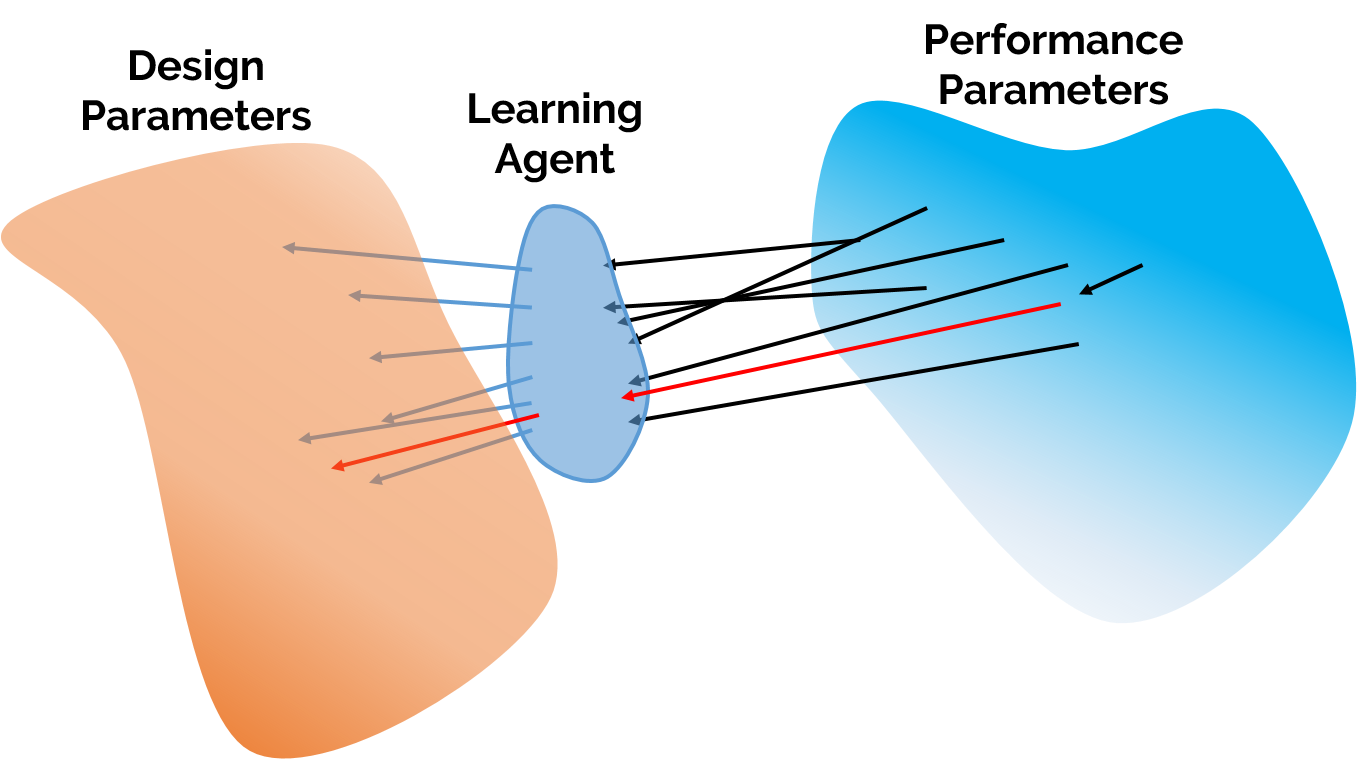}
\caption{By using our data to train a learning agent, a continuous mapping is generated allowing new designs to be computed automatically.}
\label{fig:LearningAgent}
\end{figure}

In addition to enabling a continuous representation of our design problem, using a learning agent to store data relationships in a functional form drastically reduces the amount of memory and computational overhead required to describe the function. With discrete data, estimating this function by choosing the closest discrete data-point would require thousands (or more) data-points to be stored and sorted. By instead using these data-points to fine-tune a continuous, global function we only need to store the function parameters. Then as new data is encountered, the network parameters can update but the amount of storage required will remain relatively constant.

 We represent the collective group of all our chosen performance concerns in the performance parameter space, which will be the input to our learning agent. The output of the network is the design parameter space, whose shape is entirely up to us. As designers this is the most important part in this process, as the representation we choose explicitly determines how difficult this learning process will be. Fundamentally, the representation chosen to describe the design parameter space needs to uphold two properties. First, it must be complete, i.e., every single performance parameter we wish to design must be possible to reach from the design parameter space. Without this property, our agent might not be physically capable of describing the system we are trying to find. Secondly, it should be compact. As the dimensionality of our representation grows larger, the complexity of the problem explodes exponentially \cite{Bellman:1957}. By choosing a representation that is compact, we can maximize the potential and efficiency of our learning agent. 

For single freeform surface designs we propose the use of spherical harmonics to describe the design parameter space. As discussed in \cite{Gannon:18}, the spherical harmonics form an orthonormal basis on top of the spherical emission profile from most light sources. Not only does this guarantee completeness since the spherical harmonics form a basis, it also guarantees compactness due to the orthonormality since every change in basis functions is guaranteed to have a uniquely meaningful contribution to the surface shape. 

\section{Examples}
\begin{figure}[!htb]
\centering
\includegraphics[width=\linewidth]{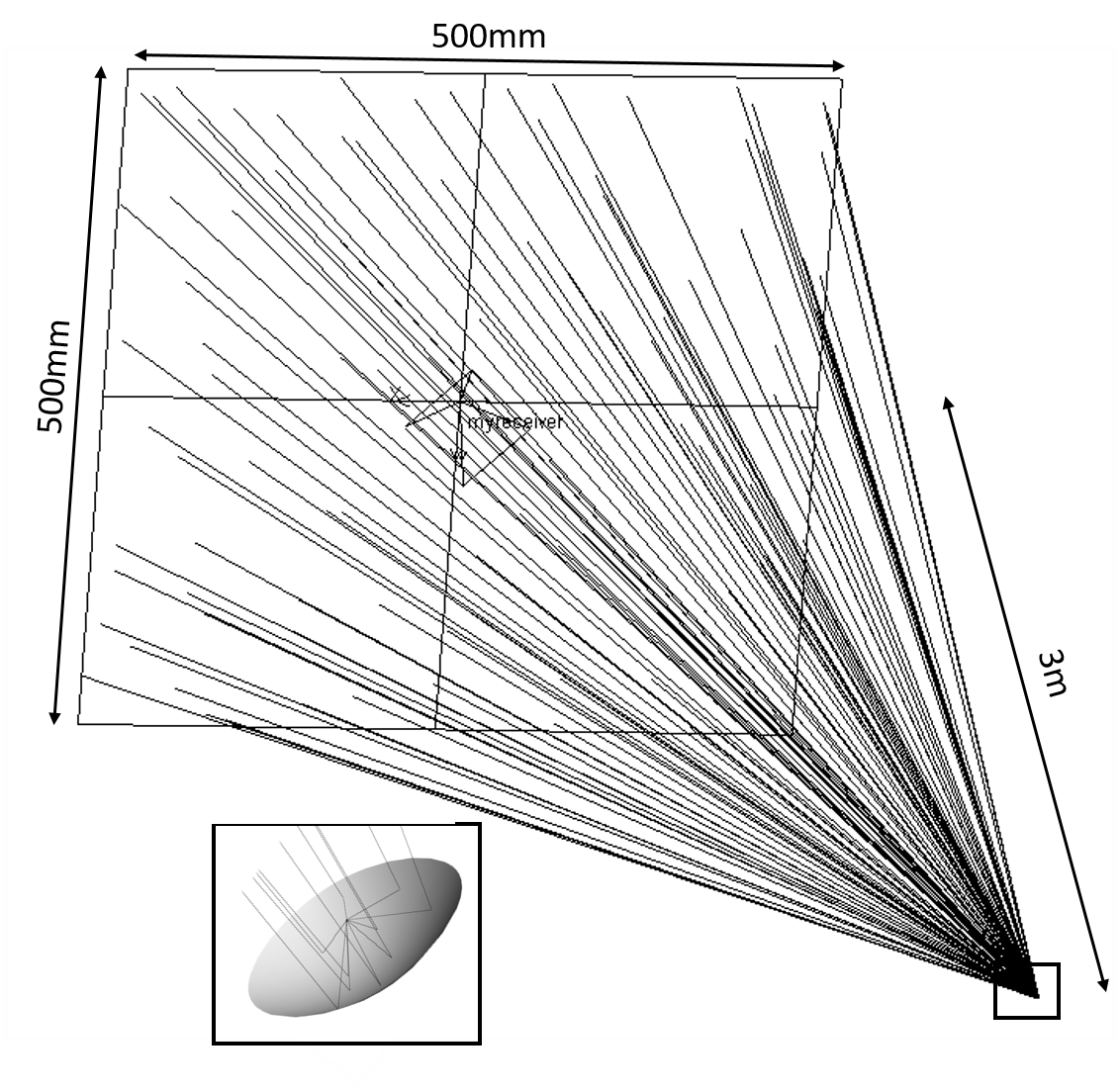}
\caption{A demonstration of the reflector geometry and target setup created in LightTools.}
\label{fig:geometry1}
\end{figure}

\begin{figure}[!htb]
\centering
\includegraphics[width=\linewidth]{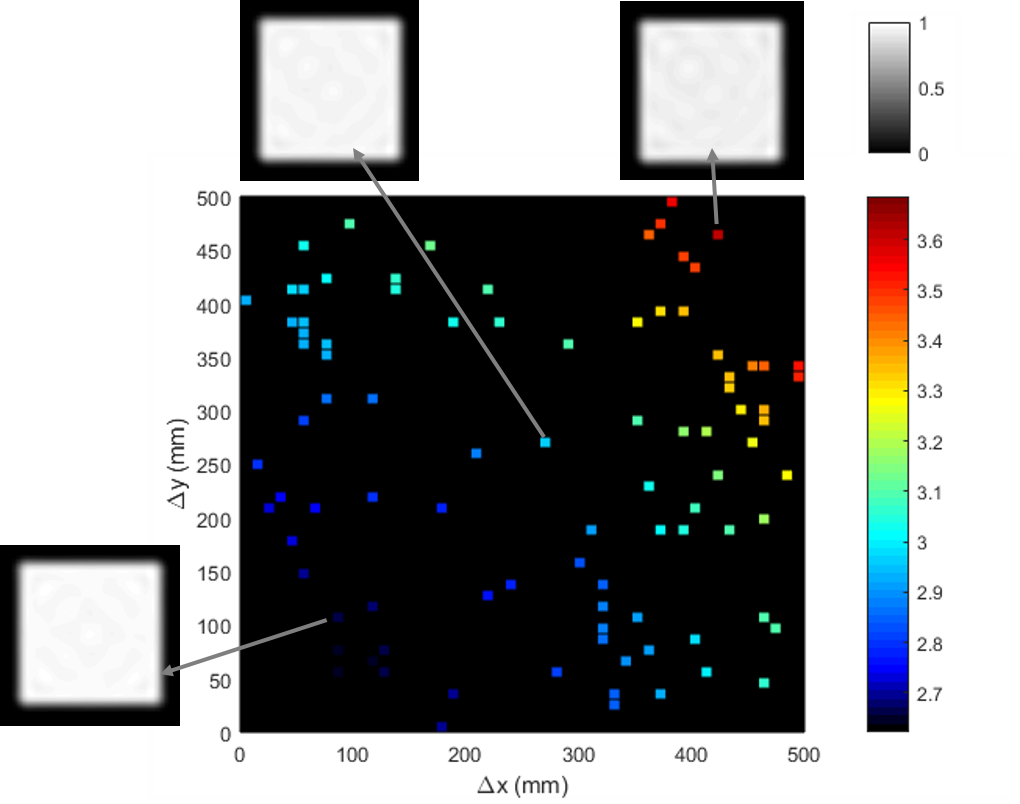}
\caption{A visualization of the database generated for training the neural network. The x and y axes correspond to the x and y offset of the target. The colormap indicates the non-uniformity of the generated illumination pattern, as calculated by $100\%(RMSdeviation/mean)$ on the entire 81x81 grid with a 5 pixel smoothing kernel to reduce statistical error from the raytracing. Black regions indicate no data present. The small figures show the illumination patterns produced by the indicated lenses, shown in linear scale.}
\label{fig:DatabaseExample}
\end{figure}

As an initial demonstration of this approach, we looked into the design of uniform square patterns at a specified x and y offset from the optical axis. We generated a database of 100 datapoints sampled from a uniform random distribution in x and y between 0 and 500mm away from the optical axis using LightTool's freeform design toolbox through the Matlab API. After construction, each lens in this database was fit with Spherical Harmonic terms up to 10$^{th}$ order, creating a total of 121 Spherical Harmonic parameters. Additionally, a slight tilt was added to the lenses to help improve the off-axis performance. An example geometry is shown in Fig. \ref{fig:geometry1} where reflectors made out of aluminum accepting 170$^\circ$ of a lambertian point source were created to produce square patterns onto a 500mmx500mm target 3m away with the specified x and y offset, and the final database performance is shown in Fig. \ref{fig:DatabaseExample}. 

\begin{figure}[!htb]
\centering
\includegraphics[width=\linewidth]{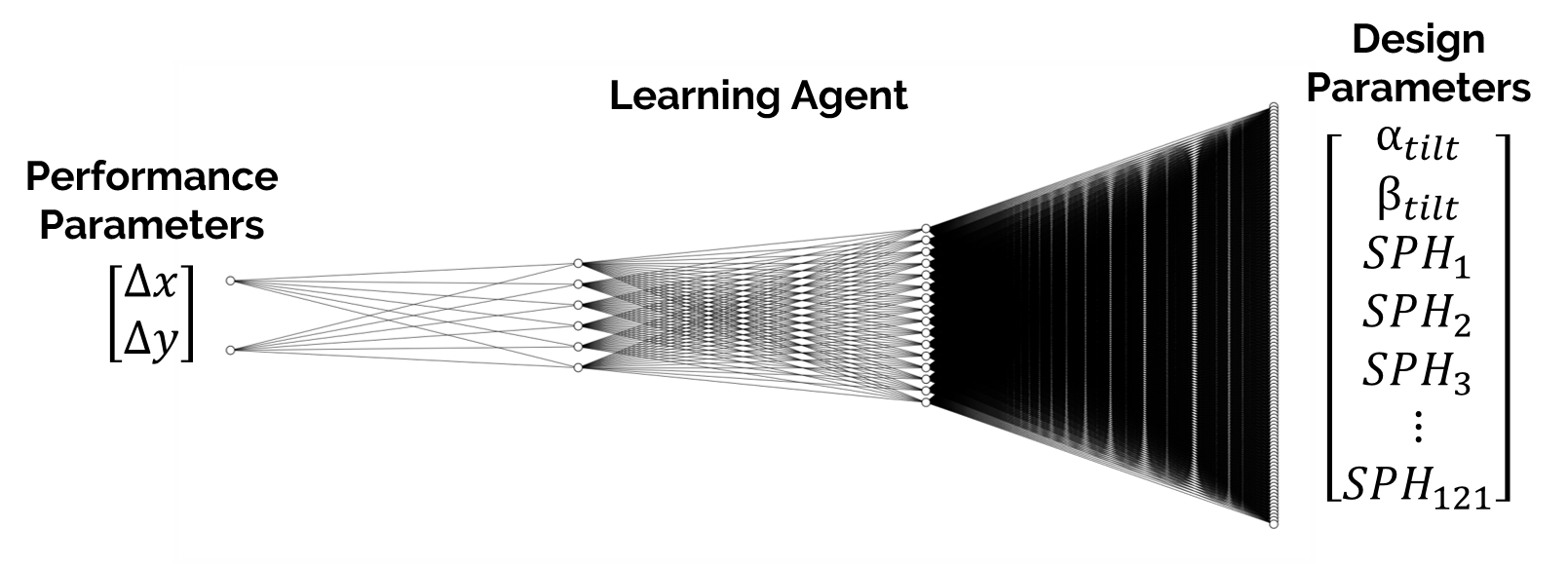}
\caption{Visualization of the artificial neural network created to design uniform squares at a defined x,y offset from the optical axis using lens surfaces described by spherical harmonics. The network had 2 input neurons, followed by two hidden layers of 6 and 16 neurons respectively, ending in an output layer with 123 neurons. The first two outputs were used to denote the $\alpha$ and $\beta$ tilt of the mirror while the rest each represented a unique spherical harmonic term}
\label{fig:squareNetwork}
\end{figure}

Together, the spherical harmonic terms and the tilt values made up the design parameter space, while the performance parameter space was described entirely by two values indicating the x and y offset of the target. A feedforward artificial neural network was used as the learning agent, with topology as shown in Fig. \ref{fig:squareNetwork}.

\begin{figure}[!htb]
\centering
\includegraphics[width=\linewidth]{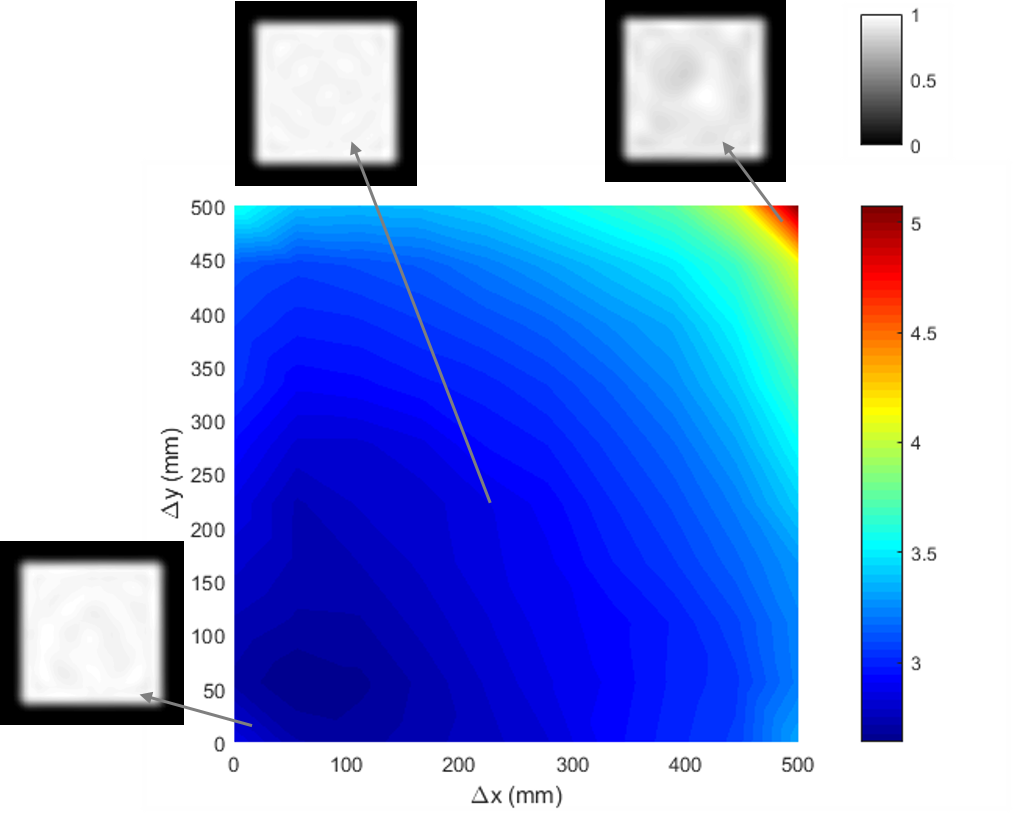}
\caption{A plot of the final network performance to be compared with the input data in Fig. \ref{fig:DatabaseExample}. The x and y axes correspond to the x and y offset of the target, while the colormap denotes the non-uniformity of the generated pattern as calculated by $100\%(RMSdeviation/mean)$ on the entire 81x81 grid with a 5 pixel smoothing kernel to reduce statistical error from the raytracing.}
\label{fig:NetworkPerformEllipse}
\end{figure}

The network was trained in Matlab's neural network toolbox using Levenberg-Marquardt back-propagation \cite{hagan1994training} with error defined as the mean squared difference between the desired tilts and spherical harmonic terms and the tilts and spherical harmonic terms output by the network. This network took about an hour to train on a single core of a 2.59 Ghz processor. In lieu of a validation dataset, the learning agent's performance was evaluated directly by inputting x and y offset values to the network, building the prescribed lenses and measuring the difference between illumination pattern produced on the target and the desired pattern. A plot of this performance is shown in Fig. \ref{fig:NetworkPerformEllipse}, where we can see that the network has successfully generalized the sparse information provided in the database to generate high-performance designs throughout the entire region. 

\section{Generality}
\begin{figure}[!htb]
\centering
\includegraphics[width=\linewidth]{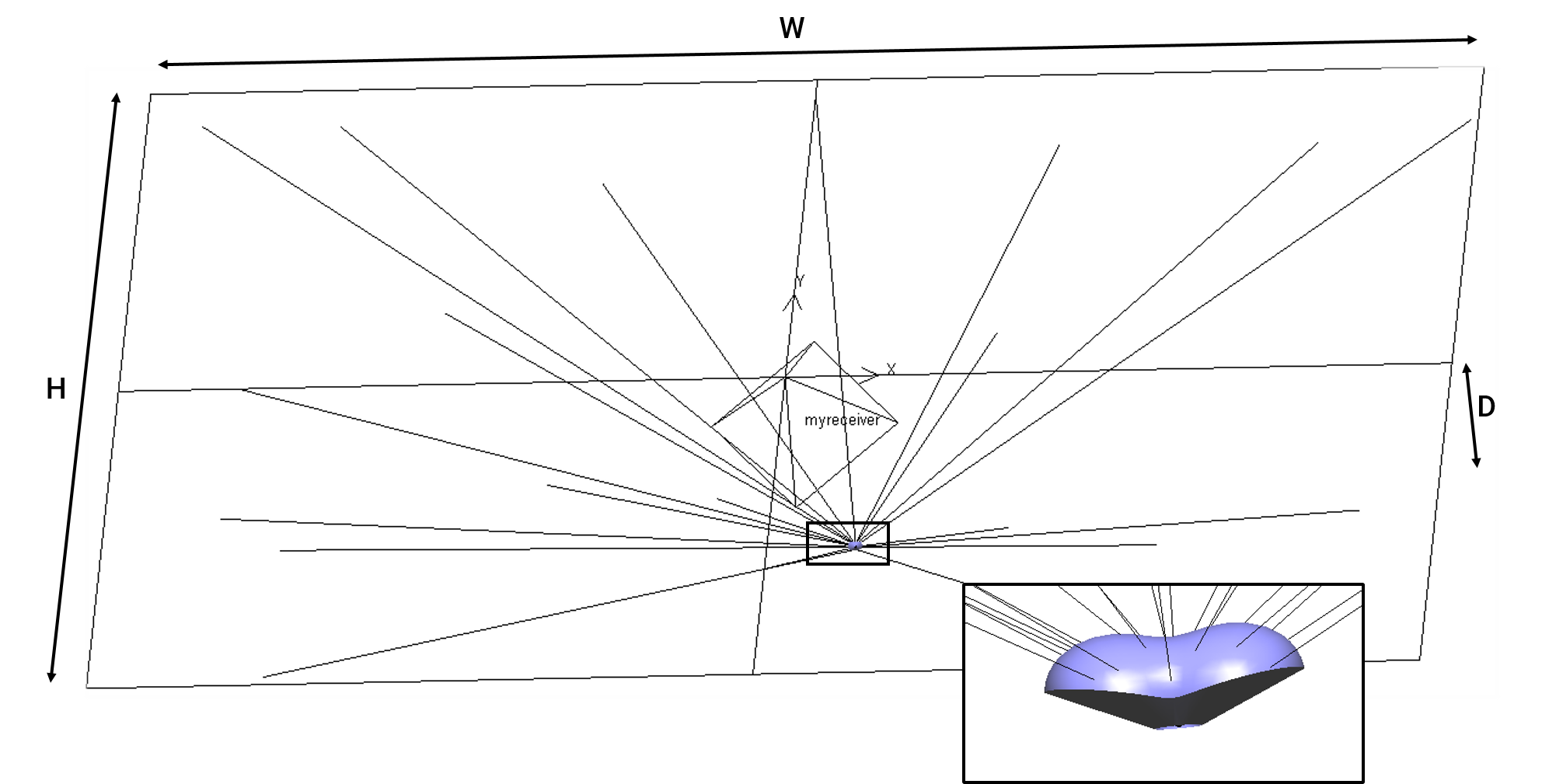}
\caption{A demonstration of the refractive geometry and target setup created in LightTools.}
\label{fig:geometry2}
\end{figure}

To demonstrate the general utility of this approach, we also generated a database of refractive lenses made from PMMA collecting a 140$^\circ$ emission of a lambertian point source to produce uniform rectangular illumination patterns with specified width and height values, at a selected distances away from a target. An example geometry is given in Fig. \ref{fig:geometry2}. 

\begin{figure}[!htb]
\centering
\includegraphics[width=\linewidth]{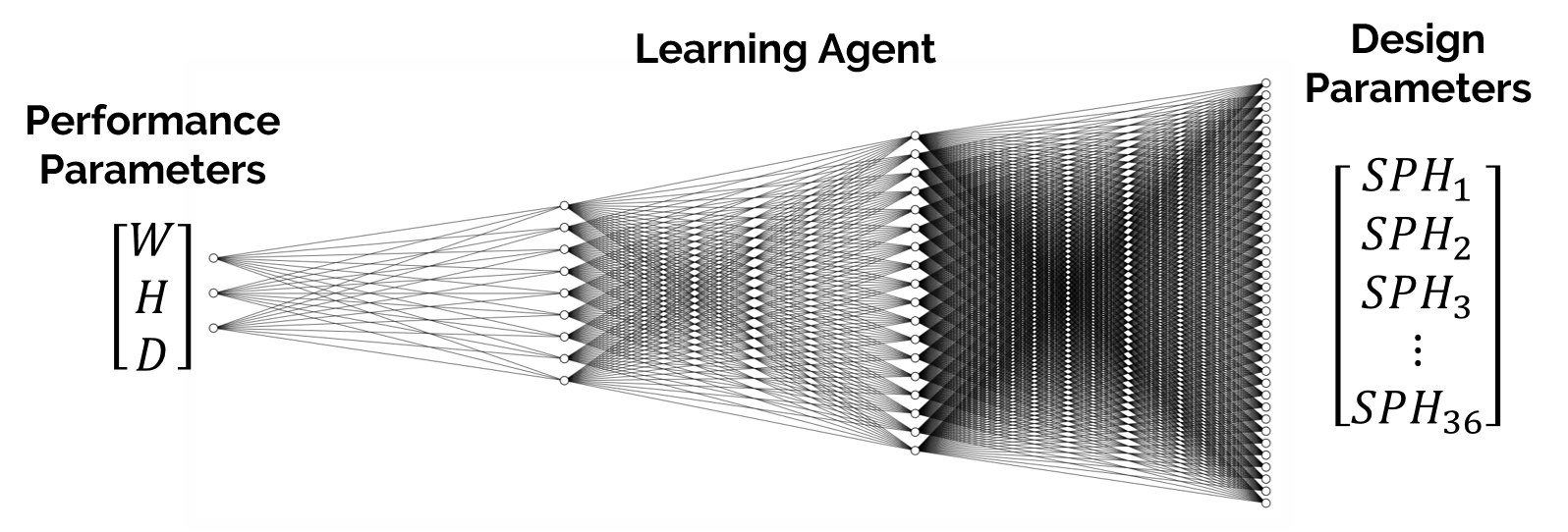}
\caption{Visualization of the artificial neural network created to design uniform rectangles with a defined height and width at a given distance away from the target using lens surfaces described by spherical harmonics. The network had 3 input neurons, followed by two hidden layers of 9 and 18 neurons respectively, ending in an output layer with 36 neurons, each corresponding to a unique spherical harmonic term.}
\label{fig:Network}
\end{figure}

This database was also made using LightTool's freeform design toolbox. Although we still require 10$^{th}$ order spherical harmonics to accommodate the rectangular target shape,  because of the quadrant symmetry of this problem we only need to use 36 spherical harmonic terms for the network output. The network topology for this problem is shown in Fig. \ref{fig:Network}

This network took around 5 minutes to train using the same processor as before. The time decrease compared to the first network is due to the significantly reduced number of output parameters and the more complete training dataset. In this case, we used a uniformly distributed dataset across the entire 3-D design space. Each dimension had 20 points in it, meaning the final database contained 800 designs. Because the performance within the training data was imperceptibly identical to the final network output, to avoid presenting two identical plots we simply show the final performance of the network generated designs below in Fig. \ref{fig:offaxisPerformance}. The training database has been included in the supplementary materials for verification. 

The performance decrease near the corners (at an aspect ratio of 2) is due to the increasingly complexity of the surface shape required to generate high aspect ratio designs. Because we are using a fixed polynomial order we are unable to completely describe the high-frequency surface components that would be required to produce the necessary surface shape causing a reduction in performance, meaning this approach of using polynomial coefficients is likely best suited for designs where symmetry can be exploited or the target distribution has few high-frequency components (which we believe is the case in many illumination design problems). 

\begin{figure}[!htb]
\centering
\includegraphics[width=\linewidth]{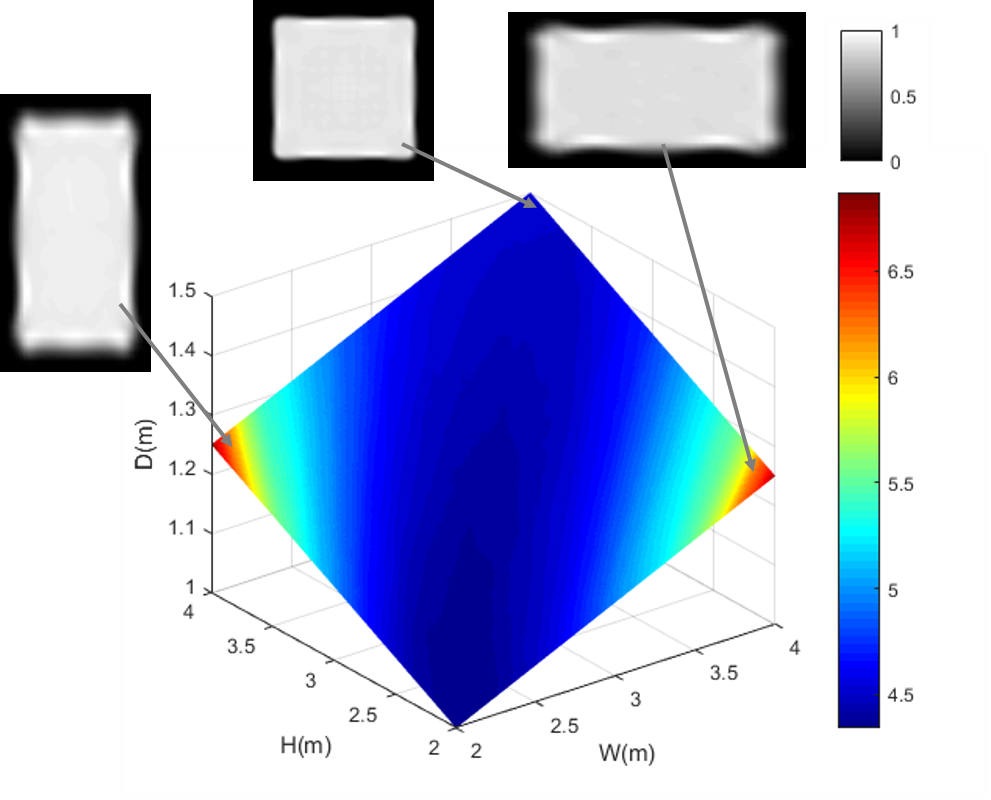}
\caption{Performance of lenses generated by the network to produce a uniform rectangular illumination pattern with a given height and width, at a distance D away. The colormap indicates the non-uniformity of the generated illumination pattern in percent, calculated as $100\%(RMSdeviation/mean)$ on a 41x41 grid with a 3 pixel smoothing kernel. The network was trained on the entire data cube but a representative diagonal slice was shown for visualization.}
\label{fig:offaxisPerformance}
\end{figure}

\begin{figure}[!htb]
\centering
\includegraphics[width=\linewidth]{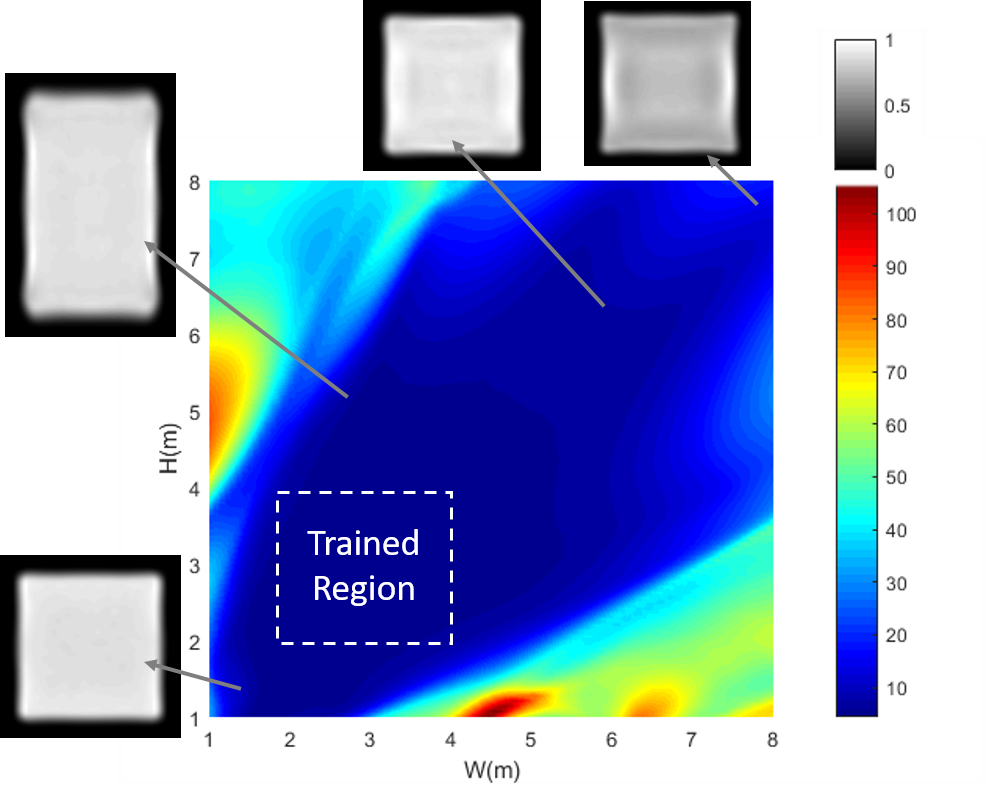}
\caption{Visualization of network performance when width and height values outside the training region were input. Distance values were held the same as in Fig. \ref{fig:offaxisPerformance}. Error was calculated as $100\%(RMSdeviation/mean)$ on a 41x41 grid with a 3 pixel smoothing kernel. }
\label{fig:generalRectangle}
\end{figure}

To further demonstrate the effectiveness of this method, we input performance parameters far outside the training dataset into the second network to see how well it could generalize its learned knowledge. In Fig. \ref{fig:generalRectangle}, we tested the network using width and height values between 1m and 8m long. Unsurprisingly, the performance is quite poor in regions with large aspect ratios where higher order polynomial terms would be needed to describe the increasingly complex surfaces. However, within the domain of surfaces that can reasonably described using 10$^{th}$ order polynomials, the network performed quite well. Despite our training dataset only having information for targets 2-4m wide, the network seems to have generalized this information to produce high quality results across a much larger region of the design space. Even at the points where the performance was degraded, the surface shape is still quite close to the ideal prescription and could serve as an excellent starting point for optimization. Once optimized, those points could then be re-taught to the network to improve future performance.

\begin{figure}[!htb]
\centering
\includegraphics[width=\linewidth]{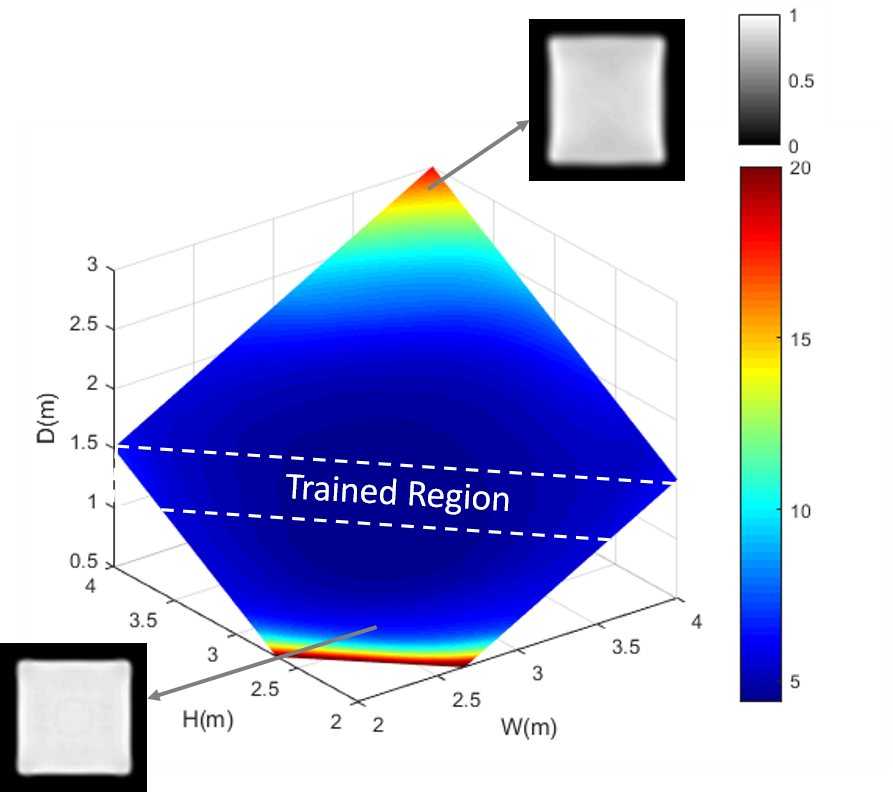}
\caption{Visualization of network performance when target distances outside the training region were input. Error was calculated as $100\%(RMSdeviation/mean)$ on a 41x41 grid with a 3 pixel smoothing kernel.}
\label{fig:generalDistance}
\end{figure}

We also investigated testing the network at different distances from the source than what it had been taught. Looking at Fig. \ref{fig:generalDistance} we can see the performance is quite good. Although our training dataset was only between 1 and 1.5m, the network produced high performing designs from 0.5 to 3m, nearly 5x the range of the training data.

Although the process of producing the lenses in both of these examples was rather time consuming (taking a couple minutes per design) using the representation in this paper we have found an extremely efficient mapping from the performance space to the design space. Using the proposed method, future lenses can be generated in milliseconds, which is around a 5-6 order of magnitude speed increase. Additionally, because of recent hardware advances in performing artificial neural network computations on GPUs and even Tensor Processing Units (TPUs), these computations can be done almost entirely in parallel meaning many thousands of designs can be generated simultaneously with little or no additional time requirements.

 While this might feel like a relatively unimportant improvement, as waiting a minute for a design might not seem too long; with a speed improvement of this magnitude the ability to continuously scroll through design options becomes a real possibility, which might be particularly useful in cases where design trade-offs need to be considered. Rather than picking a best candidate from a small number of completed designs, the designer could explore the continuum of possibilities in real-time to find a solution that best suits their needs. 

\section{Conclusion}
In this paper we demonstrate a freeform illumination design method using machine learning.  By using an artificial neural network outputting orthogonal polynomial coefficients, we are able to generalize relationships between input performance parameters and output lens shape. In this paper we created a network to perform the somewhat abstracted tasks of generating uniform squares with a desired x-y offset and uniform rectangles of a desired width and height at a given distance from the target. In doing so, we demonstrated the capability of these networks to learn design goals at a higher level than the PDE boundary equations which would be required using direct design methods and create designs with a dramatically reduced computational burden, enabling significant speed and memory reductions (the example neural networks in this paper were stored in 43kb and 19kb Matlab files, respectively). There is still plenty of work to be done in this area, but we believe speed improvements of this magnitude may open up the possibility for entirely new design approaches in the future.

\section{Funding Information}
This research was supported by the National Science Foundation Graduate Research Fellowship Grant DGE-1746060.



\bibliography{Bibliography.bbl}



\end{document}